\documentclass[11pt]{article}

\usepackage[top=1.15in, bottom=1.15in, left=1.20in, right=1.20in]{geometry}
\usepackage{times}
\usepackage[T1]{fontenc}
\usepackage[utf8]{inputenc}
\usepackage{microtype}
\usepackage{graphicx}
\usepackage{amsmath}
\usepackage{amssymb}
\usepackage{booktabs}
\usepackage{tikz}
\usetikzlibrary{arrows.meta, positioning, shapes.geometric, fit, backgrounds}
\usepackage{url}
\usepackage{enumitem}
\usepackage{titlesec}
\usepackage[colorlinks=true,
            linkcolor=blue!65!black,
            citecolor=blue!65!black,
            urlcolor=blue!65!black]{hyperref}

\titleformat{\section}{\large\bfseries}{\thesection}{0.8em}{}
\titleformat{\subsection}{\normalsize\bfseries}{\thesubsection}{0.8em}{}
\titlespacing*{\section}{0pt}{1.4ex plus 0.4ex}{0.8ex plus 0.2ex}
\titlespacing*{\subsection}{0pt}{1.0ex plus 0.3ex}{0.5ex}

\newcommand{\res}[1]{\textbf{#1}}
\newcommand{\doi}[1]{\href{https://doi.org/#1}{\nolinkurl{#1}}}

\title{\textbf{TabuLM: Morphology-Aware Tabular Pre-training\\
               for Low-Resource Languages}}

\author{%
  \begin{tabular}[t]{ccc}
    Ireddi Rakshitha & Devavarapu Yashwanth & Pierre Ntakirutimana \\[0.25em]
    \small Software Engineer & \small Software Engineer & \small Research Associate \\
    \small Barclays          & \small Barclays          & \small Carnegie Mellon University
  \end{tabular}%
}

\date{}

\begin{document}
\maketitle

\begin{center}
  \small
  \textbf{Code:}
  \href{https://github.com/TabuLM-Research/tabulm}{\texttt{github.com/TabuLM-Research/tabulm}}\\
  \quad\textbf{Data:}
  \href{https://huggingface.co/TabuLM-Research/tabulm}{\texttt{huggingface.co/TabuLM-Research/tabulm}}
\end{center}

\vspace{0.6em}

\begin{abstract}
Structured tabular data census records, agricultural surveys, administrative
reports are ubiquitous in low-resource language contexts yet completely
absent from existing pre-training pipelines.
Meanwhile, tabular language models such as TAPAS, TaBERT, and TARTE operate
exclusively on English, and low-resource NLP models such as KinyaBERT operate
exclusively on free text.
We introduce \textbf{TabuLM}, the first language model that jointly captures
(i)~the morphological richness of Kinyarwanda and (ii)~the relational structure
of tabular data.
TabuLM extends KinyaBERT's two-tier morpho-semantic architecture with additive
row, column, and cell-type embeddings, a learned per-head table-structure
attention bias injected at every layer of the sequence transformer, and two
new pre-training objectives: \textit{Masked Cell Recovery} (MCR) and
\textit{Column Type Prediction} (CTP).
We construct \textbf{TabQA-kin}, the first native Kinyarwanda table
question-answering benchmark.
TabuLM surpasses KinyaBERT, mBERT, and XLM-R baselines on TabQA-kin by
5.7--12.7 exact-match points (overall EM), achieving 62.0\% with
question-guided inference for both lookup and comparison questions.
Zero-shot GPT-4o and GPT-4o-mini both achieve 64.0\% overall, revealing a
scale-independent LLM ceiling driven by aggregation failure (25--30\%); all
fine-tuned models break through this ceiling on aggregation (TabuLM: 79.2\%,
KinyaBERT: 88.9\%), demonstrating that domain-specific tabular fine-tuning
addresses a structural gap that larger LLMs cannot overcome.

\end{abstract}

\section{Introduction}
\label{sec:intro}

Kinyarwanda, a Bantu language spoken by over 12 million people in Rwanda and
the Great Lakes region, is one of the most morphologically complex languages in
the world.
A single Kinyarwanda verb can encode subject agreement, tense, object agreement,
aspect, and directionality in one surface form: \textit{ndagukunda}
(``I love you'') bundles five morphological slots into four syllables.
KinyaBERT~\cite{nzeyimana-niyongabo-rubungo-2022-kinyabert} addressed this
challenge with a two-tier transformer: a morpheme-level encoder per word
followed by a word-level sequence encoder, achieving the best results on five
Kinyarwanda NLP benchmarks at ACL 2022.

Yet the vast majority of Kinyarwanda data that \emph{matters for governance}
is \emph{tabular}.
Rwanda's National Institute of Statistics (NISR) publishes census data,
agricultural production statistics, school enrollment records, and health
facility surveys entirely as structured tables.
The Rwanda Agriculture Board maintains yield-by-crop-by-district-by-season
tables for over 40 crop types.
None of this information is accessible to KinyaBERT or any other existing
Kinyarwanda model, because all prior pre-training pipelines consume free
text only.

On the English side, tabular language models (TAPAS, TaBERT, TURL, TABBIE,
TARTE) have demonstrated strong gains on table question answering and
table-to-text generation but every one of these models is English-only
and assumes English tokenization, making direct transfer to morphologically
rich low-resource languages impossible.

\paragraph{The problem.}
The gap is precise: no model can \emph{simultaneously} handle Kinyarwanda's
morphological complexity and the relational structure of administrative tables.
KinyaBERT understands morphology but treats tables as plain text, losing all
row--column relational signals.
TAPAS understands table structure but uses English subword tokenization that
misfires on Kinyarwanda agglutinative forms splitting
\textit{umubare w'abaturage} (``population count'') into unrecognizable fragments
that cannot match question tokens.
The consequence is that Rwanda's richest and most policy-relevant data sources
remain effectively inaccessible to any language model.

\paragraph{What we do.}
We address this gap at both the architectural and objective level.
Architecturally, we extend KinyaBERT's two-tier transformer with three
lightweight tabular embeddings (row, column, cell type) and a learned
table-structure attention bias, adding fewer than 0.1\% new parameters to the
base model.
On the data side, we collect 172 Rwandan government tables and construct
\textbf{TabQA-kin}, the first native Kinyarwanda table QA benchmark with 526
annotated question-answer pairs.
For pre-training, we introduce two new objectives: \textit{Masked Cell Recovery}
(MCR), which forces relational reasoning across rows and columns, and
\textit{Column Type Prediction} (CTP), which ties header semantics to cell
content distributions.

\paragraph{Our solution and main findings.}
The resulting model, \textbf{TabuLM}, achieves 62.0\% exact match on TabQA-kin,
outperforming all fine-tuned baselines by 5.7--12.7 EM points.
A surprising finding emerges from our LLM comparison: GPT-4o and GPT-4o-mini
both plateau at 64.0\% overall an identical ceiling driven entirely by
failure on aggregation questions (25--30\% EM).
All fine-tuned models, including TabuLM (79.2\%) and even mBERT (80.8\%), break
through this ceiling on aggregation, revealing that the bottleneck for
Kinyarwanda tabular QA is \emph{structural alignment between entity-linked
questions and table rows} a gap that fine-tuning addresses and zero-shot
LLMs cannot.

\paragraph{Contributions.}
\begin{enumerate}[leftmargin=*, label=\arabic*.]
  \item \textbf{Architecture.} TabuLM extends KinyaBERT's sequence
    transformer with additive row, column, and cell-type embeddings, and a
    learned table-structure attention bias (same-row, same-column, header
    signals) injected at every layer via the existing \texttt{attn\_bias}
    hook in the transformer implementation.

  \item \textbf{Pre-training objectives.} Two new objectives designed for
    tabular corpora \textit{Masked Cell Recovery} (MCR), which masks entire
    cells and forces reconstruction from row and column context, and
    \textit{Column Type Prediction} (CTP), which predicts whether a column
    is numeric, textual, categorical, or temporal.

  \item \textbf{Domain adaptation of morphological processing.} We extend
    KinyaBERT's existing morphological pipeline to handle numeric expressions,
    administrative entity names, and agricultural compound terms common in
    tabular data but rare in text corpora, using a BPE fallback for
    out-of-vocabulary tokens.

  \item \textbf{TabQA-kin.} The first native Kinyarwanda table
    question-answering benchmark, with 526 question-answer pairs over 31
    tables spanning census, agriculture, education, health, and
    infrastructure domains.

  \item \textbf{Generalizability.} The tabular components are fully
    language-agnostic; the Tier~1 morpho encoder is the only
    language-specific part, requiring only a morphological analyzer and
    vocabulary, making TabuLM directly applicable to other Bantu languages
    (Swahili, Kirundi) and, via BPE fallback, to any morphologically rich
    low-resource language.
\end{enumerate}

\section{Problem Formulation}
\label{sec:problem}

We formalize the task of \emph{morphologically-aware tabular language
understanding} for low-resource languages and identify the three core challenges
that motivate our design choices.

\paragraph{Setting.}
Let $T = (H, V)$ be a table with header row $H = [h_1, \ldots, h_C]$ and data
matrix $V \in \Sigma^{R \times C}$, where each cell $v_{r,c} \in \Sigma^*$ is a
string in the target language.
In Kinyarwanda, $\Sigma^*$ encompasses morphologically complex words, numerals,
administrative place names, and agricultural compound nouns.
A natural-language question $Q \in \Sigma^*$ is posed against $T$.

\paragraph{Table Question Answering.}
Following TAPAS~\cite{herzig-etal-2020-tapas}, we treat table QA as
\emph{cell selection}: the model assigns a probability to each cell and predicts:
\begin{equation}
(r^*, c^*) = \arg\max_{r,c}\; P_\theta(r,c \mid Q, T).
\end{equation}
Four question types arise in Kinyarwanda administrative data:
\textsc{lookup} (retrieve a specific cell given row and column cues),
\textsc{comparison} (identify which of two named entities has a higher/lower
value in a column),
\textsc{aggregation} (identify the entity with the global maximum/minimum in a
column), and
\textsc{count} (how many rows satisfy a numeric threshold excluded from
cell-selection evaluation since the answer is a derived number, not a cell).

\paragraph{Core challenges.}
Three challenges compound in this setting and are absent from English tabular QA:

\begin{enumerate}[leftmargin=*, label=\arabic*.]
  \item \textbf{Morphological surface variation.}
    A Kinyarwanda noun phrase in a question may appear in a different surface
    form than its counterpart in a table cell due to inflectional changes from
    noun-class agreement, locative affixes, or contextual shortening.
    Standard subword tokenizers fragment these forms unpredictably, breaking
    the stem-level match that would otherwise link question tokens to cell
    tokens.
    KinyaBERT's morphological pipeline operating at the stem and affix
    level resolves these mismatches at the encoding stage.

  \item \textbf{Absence of tabular pre-training signal.}
    No existing Kinyarwanda pre-training corpus includes structured tables.
    Language models trained on text alone have never seen the relational
    patterns governing tabular reasoning: that values in the same column are
    comparable, that row identities are bound to their column header, or that
    numeric aggregation follows a predictable positional structure.
    Without explicit relational pre-training, the model must acquire these
    patterns entirely from the small fine-tuning set (420 training examples).

  \item \textbf{Low-resource data scarcity.}
    English tabular QA benchmarks (WikiTQ, SQA, HiTab) contain tens of
    thousands of annotated pairs; our entire Kinyarwanda corpus spans 172
    tables and 526 QA pairs.
    Pre-training must be \emph{sample-efficient}, and the tabular architecture
    must transfer structural knowledge from pre-training to fine-tuning without
    overfitting on an extremely small training set.
\end{enumerate}

TabuLM directly targets all three: KinyaBERT's Tier~1 morpho encoder addresses
(1); the MCR and CTP objectives with tabular embeddings address (2); and
warm-starting from KinyaBERT with parameter-efficient tabular additions addresses
(3).

\section{Background: KinyaBERT}
\label{sec:background}

KinyaBERT~\cite{nzeyimana-niyongabo-rubungo-2022-kinyabert} is the
state-of-the-art pre-trained language model for Kinyarwanda.
Its architecture consists of two stacked transformers operating at different
granularities.

\paragraph{Tier 1 Morpho Transformer.}
A 4-layer, 4-head transformer processes each word individually.
Its input is a sequence of per-word morphological embeddings:
$[\mathbf{e}^{\text{pos}}_1, \mathbf{e}^{\text{pos}}_2, \mathbf{e}^{\text{stem}}, \mathbf{e}^{\text{afset}}, \mathbf{e}^{\text{afx}}_1, \dots, \mathbf{e}^{\text{afx}}_m]$,
where $\mathbf{e}^{\text{pos}}$ are POS tag embeddings,
$\mathbf{e}^{\text{stem}}$ is the stem embedding (vocabulary prefixed by word
class: \texttt{N:}, \texttt{V:}, \texttt{NP:}), $\mathbf{e}^{\text{afset}}$
is an affix-combination signature embedding (from a vocabulary of 34,008 affix
sets), and each $\mathbf{e}^{\text{afx}}_i$ is an individual morpheme embedding.
The transformer's output vector at position 0 (the POS slot) serves as the
128-dim morphological summary $\mathbf{m}_w$ for word $w$.

\paragraph{Tier 2 Sequence Transformer.}
A 12-layer, 8-head transformer processes the sentence at word level.
Each word's input is $[\mathbf{m}_w \;\|\; \mathbf{s}_w]$, where
$\mathbf{m}_w \in \mathbb{R}^{256}$ is the projected morphological summary and
$\mathbf{s}_w \in \mathbb{R}^{256}$ is the word's stem embedding, giving a
512-dimensional token representation.
Position bias combines TUPE~\cite{ke-etal-2021-rethinking} relative position
bucketing with a POS-aware relative position bias.
Pre-training objectives are word-level masked stem prediction (NLL), affix-set
prediction (NLL), and affix distribution prediction (KL divergence).

\section{Related Work}
\label{sec:related}

\paragraph{Tabular language models.}
TAPAS~\cite{herzig-etal-2020-tapas} extends BERT with positional embeddings for
rows and columns, enabling table QA via cell selection.
TaBERT~\cite{yin-etal-2020-tabert} jointly encodes utterances and tables via
linearized ``content snapshots''.
TURL~\cite{deng-etal-2022-turl} focuses on entity linking and relation
extraction from Wikipedia tables.
TABBIE~\cite{iida-etal-2021-tabbie} uses independent column and row encoders
with masked cell prediction.
Kim et al.~\cite{kim-etal-2025-tablefm} propose a table foundation model
using knowledge pre-training on structured tabular data for downstream
prediction tasks.
TEmBed~\cite{burak-etal-2026-tembed} introduces a universal tabular embedding
benchmark and explicitly notes the multilingual gap.
\emph{All of these models are English-only and assume whitespace tokenization.}

\paragraph{Low-resource NLP for African languages.}
KinyaBERT~\cite{nzeyimana-niyongabo-rubungo-2022-kinyabert} and its retrieval
extension KinyaColBERT~\cite{nzeyimana-2023-kinyacolbert} address Kinyarwanda
free text only.
TaTA~\cite{gehrmann-etal-2023-tata} is a multilingual table-to-text dataset for
four African languages (Hausa, Igbo, Swahili, Yoruba) but not Kinyarwanda, and
it is an evaluation resource rather than a pre-training model.
m3TQA~\cite{anonymous-2025-m3tqa} is a 97-language table QA benchmark including
Kinyarwanda, but tables are LLM-translated from Chinese and no pre-trained
representations are released.
\emph{No prior work pre-trains on structured tabular data for any low-resource
African or Asian language.}

\paragraph{Morphologically-aware NLP.}
Beyond KinyaBERT, morphologically aware approaches include
CanineFormer~\cite{clark-etal-2022-canine}, character-level models, and various
subword segmentation schemes~\cite{rust-etal-2021-good}.
None incorporate tabular structure.

\section{TabuLM}
\label{sec:model}

\subsection{Table Serialization}
\label{sec:serialization}

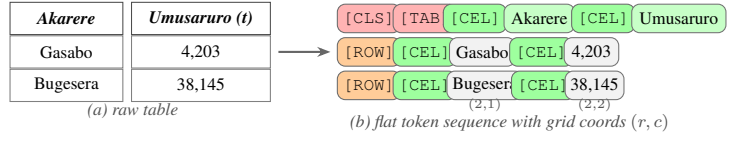
\begin{figure}[t]
\centering
\begin{tikzpicture}[
  scale=0.82, every node/.style={transform shape},
  thdr/.style ={draw=black!70, fill=white, minimum height=0.52cm,
                font=\scriptsize\bfseries\itshape, align=center, inner sep=3pt},
  tcell/.style={draw=black!50, fill=white, minimum height=0.52cm,
                font=\scriptsize, align=center, inner sep=3pt},
  tok/.style  ={draw=black!58, fill=#1, rounded corners=3pt,
                minimum height=0.46cm, font=\scriptsize\ttfamily,
                align=center, inner sep=2.5pt},
  ctok/.style ={draw=black!55, fill=#1, rounded corners=3pt,
                minimum height=0.46cm, font=\scriptsize,
                align=center, inner sep=2.5pt},
  arr/.style  ={-{Stealth[length=5pt]}, thick, black!60}
]

\node[thdr,  minimum width=1.82cm] (h1)  at (0.00,  0.90) {Akarere};
\node[thdr,  minimum width=2.24cm] (h2)  at (2.17,  0.90) {Umusaruro (t)};
\node[tcell, minimum width=1.82cm] (v11) at (0.00,  0.36) {Gasabo};
\node[tcell, minimum width=2.24cm] (v12) at (2.17,  0.36) {4{,}203};
\node[tcell, minimum width=1.82cm] (v21) at (0.00, -0.18) {Bugesera};
\node[tcell, minimum width=2.24cm] (v22) at (2.17, -0.18) {38{,}145};
\node[font=\scriptsize\itshape, gray!65!black] at (1.07, -0.58) {(a) raw table};

\draw[arr] (3.42, 0.36) -- (4.26, 0.36);

\def\ox{4.50}

\node[tok=red!30,    minimum width=0.72cm] at (\ox+0.38,  0.90) {[CLS]};     
\node[tok=red!30,    minimum width=0.72cm] at (\ox+1.28,  0.90) {[TAB]};     
\node[tok=green!38,  minimum width=0.66cm] at (\ox+2.09,  0.90) {[CEL]};     
\node[ctok=green!22, minimum width=1.14cm] at (\ox+3.13,  0.90) {Akarere};   
\node[tok=green!38,  minimum width=0.66cm] at (\ox+4.16,  0.90) {[CEL]};     
\node[ctok=green!22, minimum width=1.52cm] at (\ox+5.38,  0.90) {Umusaruro}; 

\node[tok=orange!40, minimum width=0.72cm] at (\ox+0.38,  0.36) {[ROW]};     
\node[tok=green!38,  minimum width=0.66cm] at (\ox+1.28,  0.36) {[CEL]};     
\node[ctok=gray!10,  minimum width=1.06cm] at (\ox+2.21,  0.36) {Gasabo};    
\node[tok=green!38,  minimum width=0.66cm] at (\ox+3.16,  0.36) {[CEL]};     
\node[ctok=gray!10,  minimum width=0.88cm] at (\ox+3.97,  0.36) {4,203};     

\node[tok=orange!40, minimum width=0.72cm] at (\ox+0.38, -0.18) {[ROW]};     
\node[tok=green!38,  minimum width=0.66cm] at (\ox+1.28, -0.18) {[CEL]};     
\node[ctok=gray!10,  minimum width=1.14cm] at (\ox+2.25, -0.18) {Bugesera};  
\node[tok=green!38,  minimum width=0.66cm] at (\ox+3.19, -0.18) {[CEL]};     
\node[ctok=gray!10,  minimum width=0.92cm] at (\ox+4.02, -0.18) {38,145};    

\node[font=\tiny, gray!65!black] at (\ox+2.25, -0.50) {$(2,\!1)$};
\node[font=\tiny, gray!65!black] at (\ox+4.02, -0.50) {$(2,\!2)$};

\node[font=\scriptsize\itshape, gray!60!black] at (\ox+2.65, -0.78)
     {(b) flat token sequence with grid coords $(r, c)$};

\end{tikzpicture}
\caption{Table serialization. A two-row Kinyarwanda agricultural table (a) is
linearised into a flat token sequence (b). Red \texttt{[CLS]/[TAB]} open the
table; orange \texttt{[ROW]} separates rows; green \texttt{[CEL]} precedes
each cell. Grid coordinates $(r,c)$ are attached to every token and feed the
row, column, and cell-type embedding matrices as well as the MCR and CTP
objectives.}
\label{fig:serialization}
\end{figure}

We serialize a table with $R$ data rows and $C$ columns as a flat token sequence:
\begin{align*}
&\texttt{[CLS]}\;\texttt{[TAB]}\;\texttt{[CEL]}\;h_1\;\texttt{[CEL]}\;h_2\;\cdots \\
&\texttt{[ROW]}\;\texttt{[CEL]}\;v_{1,1}\;\texttt{[CEL]}\;v_{1,2}\;\cdots \\
&\texttt{[ROW]}\;\texttt{[CEL]}\;v_{2,1}\;\cdots
\end{align*}
where $h_c$ is the header of column $c$, $v_{r,c}$ is the cell value at row $r$,
column $c$, and special tokens \texttt{[TAB]}, \texttt{[ROW]}, \texttt{[CEL]}
mark structural boundaries.
Fig.~\ref{fig:serialization} illustrates this serialization on a concrete
Kinyarwanda table.

Each content word is morphologically analyzed (or BPE-tokenized as fallback)
by the same pipeline as KinyaBERT.
Every token is annotated with its \emph{grid coordinates} $(r, c)$ and
\emph{cell type} $t \in \{\textsc{header}, \textsc{numeric}, \textsc{text}, \textsc{categorical}, \textsc{date}\}$,
detected by lightweight regex rules.

\subsection{Tabular Embeddings}
\label{sec:embeddings}

The sequence transformer's input for each token is augmented with three
additive embeddings (same dimensionality $d=512$ as the morpho-stem
representation):
\begin{equation}
\mathbf{x}_i = \underbrace{\mathbf{m}_i \,\|\, \mathbf{s}_i}_{\text{KinyaBERT}} + \underbrace{\mathbf{E}^R_{r_i} + \mathbf{E}^C_{c_i} + \mathbf{E}^T_{t_i}}_{\text{TabuLM addition}}
\end{equation}
where $\mathbf{E}^R \in \mathbb{R}^{R_{\max} \times d}$,
$\mathbf{E}^C \in \mathbb{R}^{C_{\max} \times d}$, and
$\mathbf{E}^T \in \mathbb{R}^{5 \times d}$ are learned embedding matrices for
rows, columns, and cell types respectively, with $R_{\max}=64$ and $C_{\max}=24$.
Padding index 0 is reserved for special tokens (\texttt{[CLS]},
\texttt{[TAB]}, \texttt{[ROW]}).
This additive design is consistent with how standard BERT incorporates absolute
position embeddings, requires no change to the transformer architecture, and
keeps the model dimension unchanged at $d=512$.

\subsection{Table-Structure Attention Bias}
\label{sec:attn_bias}

We augment the pre-softmax attention weights at every layer with a learned
table-structure bias.
For tokens $i$ and $j$ in the same batch, with row indices $r_i, r_j$ and
column indices $c_i, c_j$:
\begin{equation}
b^h_{ij} = \beta^h_R \cdot \mathbf{1}[r_i = r_j]
           + \beta^h_C \cdot \mathbf{1}[c_i = c_j]
           + \beta^h_H \cdot \mathbf{1}[r_i=1 \lor r_j=1]
\end{equation}
where $\beta^h_R, \beta^h_C, \beta^h_H \in \mathbb{R}$ are learned scalars per
attention head $h$ (total $3H=24$ new parameters, $H=8$).
The final attention bias is:
\begin{equation}
\hat{\mathbf{A}}^h = \mathbf{A}^h_{\text{pos}} + \mathbf{B}^h_{\text{table}}
\end{equation}
where $\mathbf{A}^h_{\text{pos}}$ is KinyaBERT's existing TUPE + POS-aware
position bias.
This bias is injected at the \texttt{attn\_bias} parameter already present in
KinyaBERT's custom transformer implementation, requiring \emph{no architectural
changes} to the transformer itself.
The three terms encode: (1)~same-row attention bonus for co-referent cells in a
record, (2)~same-column attention bonus for value comparison across rows, and
(3)~header-awareness so every token attends strongly to its column's label.

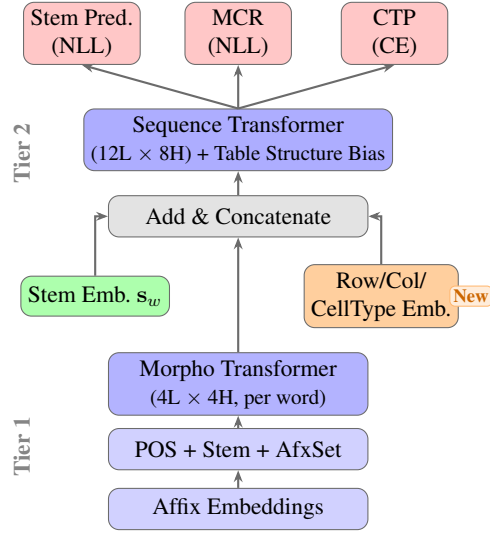
\begin{figure}[t]
\centering
\begin{tikzpicture}[
  block/.style={rectangle, rounded corners=4pt, draw=black!58, fill=#1,
                minimum width=3.90cm, minimum height=0.62cm,
                font=\small, align=center, inner sep=3pt},
  sblock/.style={rectangle, rounded corners=4pt, draw=black!58, fill=#1,
                 minimum width=1.80cm, minimum height=0.60cm,
                 font=\small, align=center, inner sep=3pt},
  hblock/.style={rectangle, rounded corners=4pt, draw=black!58, fill=#1,
                 minimum width=1.60cm, minimum height=0.62cm,
                 font=\small, align=center, inner sep=3pt},
  arrow/.style={-{Stealth[length=4.5pt]}, thick, black!58},
  scale=0.88, every node/.style={transform shape}
]

\node[block=blue!18]  (afx)    at (0, 0.00) {Affix Embeddings};
\node[block=blue!18]  (pos)    at (0, 0.90) {POS + Stem + AfxSet};
\node[block=blue!30]  (morpho) at (0, 1.88) {Morpho Transformer\\
                                              \footnotesize(4L $\times$ 4H, per word)};
\draw[arrow] (afx) -- (pos);
\draw[arrow] (pos) -- (morpho);

\node[sblock=green!32]  (stem2)  at (-2.15, 3.20) {Stem Emb.\ $\mathbf{s}_w$};
\node[sblock=orange!38] (rowcol) at ( 2.15, 3.20) {Row/Col/\\CellType Emb.};
\node[font=\scriptsize\bfseries, text=orange!80!black,
      fill=orange!14, draw=orange!55, rounded corners=2pt, inner sep=1.5pt]
      at (3.50, 3.20) {\textbf{New}};

\node[block=gray!22]  (concat) at (0, 4.40) {Add \& Concatenate};

\draw[arrow] (morpho.north) -- (concat.south);
\draw[arrow] (stem2.north)  |- (concat.west);
\draw[arrow] (rowcol.north) |- (concat.east);

\node[block=blue!32]  (seq) at (0, 5.55)
    {Sequence Transformer\\
     \footnotesize(12L $\times$ 8H) + Table Structure Bias};
\draw[arrow] (concat) -- (seq);

\node[hblock=red!22] (h1) at (-2.35, 7.15) {Stem Pred.\\(NLL)};
\node[hblock=red!22] (h2) at ( 0.00, 7.15) {MCR\\(NLL)};
\node[hblock=red!22] (h3) at ( 2.35, 7.15) {CTP\\(CE)};
\draw[arrow] (seq.north) -- (h1.south);
\draw[arrow] (seq.north) -- (h2.south);
\draw[arrow] (seq.north) -- (h3.south);

\node[font=\small\bfseries, text=black!48, rotate=90] at (-3.25, 0.94) {Tier 1};
\node[font=\small\bfseries, text=black!48, rotate=90] at (-3.25, 5.35) {Tier 2};

\end{tikzpicture}
\caption{TabuLM architecture. \textcolor{orange!80!black}{\textbf{Orange}}
components are new relative to KinyaBERT. Three inputs converge at
\emph{Add \& Concatenate}: the morphological summary $\mathbf{m}_w$ (from
Tier~1), the stem embedding $\mathbf{s}_w$, and the tabular embeddings. The
table-structure bias is injected inside the Sequence Transformer at every
layer.}
\label{fig:architecture}
\end{figure}

Fig.~\ref{fig:architecture} illustrates the full TabuLM architecture.

\subsection{Pre-training Objectives}
\label{sec:objectives}

TabuLM retains all three KinyaBERT pre-training objectives and adds two new ones:

\paragraph{(O1) Masked Stem Prediction (MSP).}
15\% of tokens are masked at word level; the model predicts the stem index
(NLL loss). Inherited from KinyaBERT.

\paragraph{(O2) Affix Set Prediction (ASP).}
Predicts the complete morphological signature (one of 34,008 affix-set classes)
for masked tokens. Inherited.

\paragraph{(O3) Affix Distribution Prediction (ADP).}
Predicts a probability distribution over individual affixes for masked tokens
(KL divergence). Inherited.

\paragraph{(O4) Masked Cell Recovery (MCR).}
15\% of cells (identified by unique $(r,c)$ pairs) are fully masked every
token in the selected cell receives \texttt{[MSK]}.
The objective is to recover the stem sequence of masked cell tokens using
context from other cells in the same row and column.
MCR uses the same stem-prediction head as MSP (no new parameters) but forces
the model to rely on relational structure: a masked population figure must be
recoverable from the district name in the same row and from other population
figures in the same column.

\paragraph{(O5) Column Type Prediction (CTP).}
For 50\% of columns in each training table, the header token is masked.
The model predicts the column's dominant cell type
$\hat{t}_c \in \{\textsc{numeric}, \textsc{text}, \textsc{categorical}, \textsc{date}\}$
(cross-entropy loss).
A 2-layer projection head maps the masked header's hidden state to 4 classes.
CTP forces the model to learn that a column labeled
\textit{Umubare w'Abaturage} (``Population Count'') must be followed by
numeric cells, enabling zero-shot column-type inference.

The total loss is:
\begin{equation}
\mathcal{L} = \mathcal{L}_{\text{MSP}} + \mathcal{L}_{\text{ASP}} + \mathcal{L}_{\text{ADP}} + \mathcal{L}_{\text{MCR}} + \mathcal{L}_{\text{CTP}}
\end{equation}

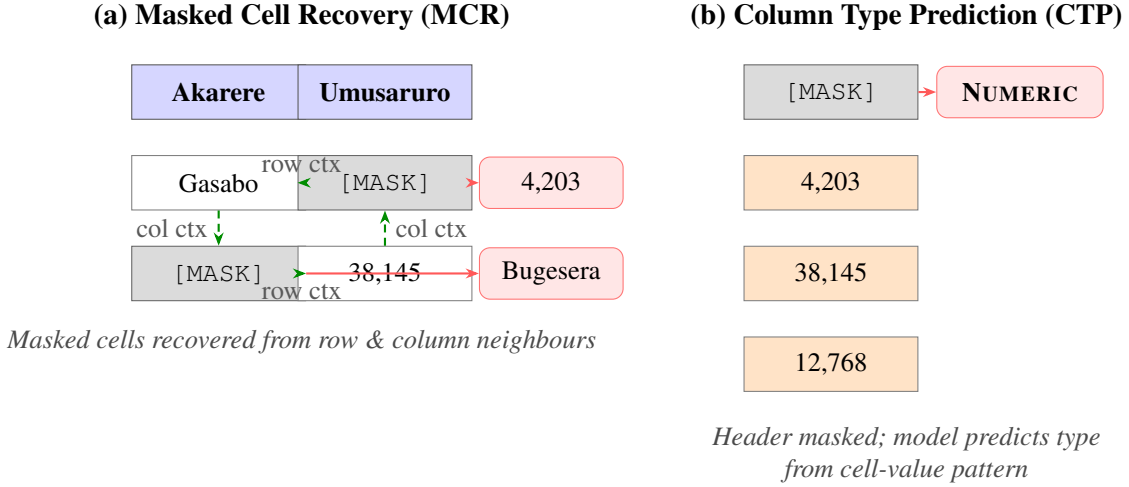
\begin{figure}[t]
\centering
\begin{tikzpicture}[
  cell/.style ={draw=black!50, fill=#1,
                minimum height=0.72cm, minimum width=2.30cm,
                font=\small, align=center, inner sep=4pt},
  hcell/.style={draw=black!65, fill=blue!16,
                minimum height=0.72cm, minimum width=2.30cm,
                font=\small\bfseries, align=center, inner sep=4pt},
  mcell/.style={draw=black!60, fill=gray!28,
                minimum height=0.72cm, minimum width=2.30cm,
                font=\small\ttfamily, align=center, inner sep=4pt},
  pred/.style ={draw=red!55, fill=red!10, rounded corners=4pt,
                minimum height=0.68cm, minimum width=1.90cm,
                font=\small, align=center, inner sep=3pt},
  carr/.style ={-{Stealth[length=5pt]}, thick, green!55!black, densely dashed},
  parr/.style ={-{Stealth[length=5pt]}, thick, red!65}
]

\node[font=\normalsize\bfseries] at (-3.70, 3.30)
     {(a)~Masked Cell Recovery (MCR)};

\node[hcell] (h1) at (-4.80, 2.30) {Akarere};
\node[hcell] (h2) at (-2.60, 2.30) {Umusaruro};

\node[cell=white] (ga)  at (-4.80, 1.10) {Gasabo};
\node[mcell]      (m12) at (-2.60, 1.10) {\texttt{[MASK]}};

\node[mcell]      (m21) at (-4.80,-0.10) {\texttt{[MASK]}};
\node[cell=white] (n22) at (-2.60,-0.10) {38,145};

\draw[carr] (ga.east)   -- node[above, font=\small, black!65] {row ctx} (m12.west);
\draw[carr] (n22.north) -- node[right, font=\small, black!65] {col ctx} (m12.south);
\draw[carr] (n22.west)  -- node[below, font=\small, black!65] {row ctx} (m21.east);
\draw[carr] (ga.south)  -- node[left,  font=\small, black!65] {col ctx} (m21.north);

\node[pred] (p12) at (-0.40, 1.10) {4,203};
\node[pred] (p21) at (-0.40,-0.10) {Bugesera};
\draw[parr] (m12.east) -- (p12.west);
\draw[parr] (m21.east) -- (p21.west);

\node[font=\small\itshape, gray!58!black] at (-3.70,-1.00)
     {Masked cells recovered from row \& column neighbours};


\node[font=\normalsize\bfseries] at (4.30, 3.30)
     {(b)~Column Type Prediction (CTP)};

\node[mcell]          (mh) at (3.30, 2.30) {\texttt{[MASK]}};
\node[cell=orange!22]      at (3.30, 1.10) {4,203};
\node[cell=orange!22]      at (3.30,-0.10) {38,145};
\node[cell=orange!22]      at (3.30,-1.30) {12,768};

\node[pred, minimum width=2.20cm, font=\small\bfseries] (tp) at (5.80, 2.30)
     {\textsc{Numeric}};
\draw[parr] (mh.east) -- (tp.west);

\node[font=\small\itshape, gray!58!black, align=center] at (4.30,-2.50)
     {Header masked; model predicts type\\from cell-value pattern};

\end{tikzpicture}
\caption{TabuLM pre-training objectives.
  \textbf{(a)}~\textit{Masked Cell Recovery} (MCR): entire cells are masked (gray);
  the model reconstructs cell content using same-row context (row identity) and
  same-column context (distributional signal) via the tabular embeddings and
  attention bias.
  \textbf{(b)}~\textit{Column Type Prediction} (CTP): a column header is masked;
  the model predicts the column's semantic type (\textsc{Numeric}, \textsc{Text},
  \textsc{Categorical}, or \textsc{Date}) from the observed cell values.}
\label{fig:objectives}
\end{figure}

\section{Tabular Data Collection}
\label{sec:data}

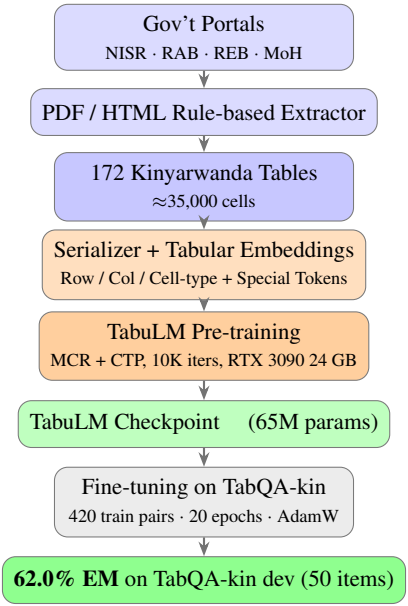
\begin{figure}[t]
\centering
\begin{tikzpicture}[
  scale=0.86, every node/.style={transform shape},
  box/.style={rectangle, rounded corners=5pt, draw=black!52, fill=#1,
              minimum width=4.60cm, minimum height=0.72cm,
              font=\small, align=center, inner sep=5pt},
  arr/.style={-{Stealth[length=5.5pt]}, thick, black!55}
]

\node[box=blue!14]    (src)  at (0,  5.70)
    {Gov't Portals\\{\scriptsize NISR $\cdot$ RAB $\cdot$ REB $\cdot$ MoH}};

\node[box=blue!14]    (ext)  at (0,  4.55)
    {PDF / HTML Rule-based Extractor};

\node[box=blue!22]    (tab)  at (0,  3.40)
    {172 Kinyarwanda Tables\\{\scriptsize $\approx$35,000 cells}};

\node[box=orange!25]  (ser)  at (0,  2.20)
    {Serializer + Tabular Embeddings\\
     {\scriptsize Row / Col / Cell-type + Special Tokens}};

\node[box=orange!38]  (pre)  at (0,  0.95)
    {TabuLM Pre-training\\
     {\scriptsize MCR + CTP, 10K iters, RTX 3090 24 GB}};

\node[box=green!25]   (ckpt) at (0, -0.25)
    {TabuLM Checkpoint \quad (65M params)};

\node[box=gray!15]    (ft)   at (0, -1.45)
    {Fine-tuning on TabQA-kin\\
     {\scriptsize 420 train pairs $\cdot$ 20 epochs $\cdot$ AdamW}};

\node[box=green!44]   (res)  at (0, -2.65)
    {\textbf{62.0\% EM} on TabQA-kin dev (50 items)};

\draw[arr] (src.south)  -- (ext.north);
\draw[arr] (ext.south)  -- (tab.north);
\draw[arr] (tab.south)  -- (ser.north);
\draw[arr] (ser.south)  -- (pre.north);
\draw[arr] (pre.south)  -- (ckpt.north);
\draw[arr] (ckpt.south) -- (ft.north);
\draw[arr] (ft.south)   -- (res.north);

\end{tikzpicture}
\caption{TabuLM data-collection and training pipeline. Starting from Rwandan
government open-data portals, tables are extracted, serialized with structural
tokens, and used to pre-train TabuLM on the MCR and CTP objectives. The
resulting checkpoint is fine-tuned on TabQA-kin to produce the final results.}
\label{fig:pipeline}
\end{figure}

Fig.~\ref{fig:pipeline} summarises the end-to-end data collection and training
pipeline described in this section.

\paragraph{Sources.}
We collect structured Kinyarwanda tables from four primary sources:
(1)~Rwanda National Institute of Statistics (NISR) open-data portal ---
census, demographic, and economic tables;
(2)~Rwanda Agriculture Board seasonal crop yield reports by district
(2010--2024);
(3)~Rwanda Education Board school enrollment and examination result tables;
(4)~Ministry of Health Rwanda health facility and disease surveillance tables.

\paragraph{Processing.}
All tables are parsed from PDF and HTML formats using rule-based extractors.
Column headers are verified to be in Kinyarwanda.
Tables with fewer than 3 rows or 2 columns, or with more than 50\% empty cells,
are discarded.
The final pre-training corpus contains 172 tables ($\sim$35,000 cells,
$\sim$170 training sequences per loop after serialization and packing to
512 tokens).

\paragraph{TabQA-kin benchmark.}
We construct the first native Kinyarwanda table QA benchmark using templated
question generation.
For each table, we instantiate Kinyarwanda question templates for four question
types: \textsc{lookup} (direct cell retrieval given a row entity and column
header), \textsc{comparison} (which row has the highest/lowest value in a
column), \textsc{count} (how many rows satisfy a numeric threshold), and
\textsc{aggregation} (column sum/average across all data rows).
Templates were authored by a native Kinyarwanda speaker and instantiated
automatically from table metadata; a second speaker independently verified
grammaticality on a random 10\% sample.
The benchmark contains 526 question-answer pairs across 31 tables, split 420/106
for train/dev (seed 42).
Count-type questions (93 items, 17.7\%) require a derived numeric answer not
present as any cell value and are excluded from cell-selection evaluation; all
reported EM scores target the remaining 433 items (186 lookup, 123 comparison,
124 aggregation).
In practice, the TabuLM dev evaluation covers 50 of the 106 dev items:
$\approx$19 count questions are excluded, and a further $\approx$37 items are
skipped at runtime due to missing CSV files, unresolvable gold cells, or
truncation beyond the 512-token limit.
Table~\ref{tab:data_stats} summarizes corpus statistics.

\begin{table}[t]
\centering
\small
\begin{tabular}{lrr}
\toprule
\textbf{Resource} & \textbf{Count} & \textbf{Source} \\
\midrule
Pre-training tables     & 172  & NISR, RAB, REB, MoH \\
Pre-training cells      & $\sim$35,000  &  \\
Pre-training sequences  & $\sim$170/loop &  \\
\midrule
TabQA-kin total QA pairs & 526 & Template generation \\
\quad \textsc{lookup}       & 186 & \\
\quad \textsc{comparison}   & 123 & \\
\quad \textsc{count}        & 93  & (excluded from eval) \\
\quad \textsc{aggregation}  & 124 & \\
Train / Dev split       & 420 / 106 & seed=42 \\
\midrule
Kinyarwanda text (mC4)  & 3.2M sent. & \cite{raffel-etal-2020-exploring} \\
\bottomrule
\end{tabular}
\caption{Dataset statistics for TabuLM pre-training corpus and TabQA-kin benchmark.}
\label{tab:data_stats}
\end{table}
\newpage
\section{Experiments}
\label{sec:experiments}

\subsection{Training Setup}
\label{sec:setup}

Table~\ref{tab:hyperparams} lists all hyperparameters; here we justify the key
choices.

\paragraph{Pre-training optimizer.}
We use LAMB~\cite{you-etal-2020-large} with peak learning rate $4\times10^{-4}$,
linear warmup over 500 steps, and linear decay to zero.
LAMB is preferred over AdamW for BERT-scale pre-training because it normalizes
parameter update norms \emph{per layer}, enabling stable training at large
effective batch sizes without manually tuning per-layer learning rates.
The $4\times10^{-4}$ peak LR follows the KinyaBERT pre-training schedule scaled
to our effective batch of 64.
Weight decay $0.01$ is applied to all non-bias, non-LayerNorm parameters,
following standard practice for transformer pre-training.

\paragraph{Effective batch size and gradient accumulation.}
We use physical batch size 8 with gradient accumulation over 8 steps (effective
batch 64).
At sequence length 512 with the full two-tier model on a single RTX~3090
(24~GB), batch~8 is the largest that fits without gradient checkpointing.
Gradient accumulation simulates the effective batch that LAMB's per-layer norm
estimates require to be statistically stable.

\paragraph{Number of iterations.}
10,000 optimizer iterations with $\sim$170 training sequences per loop yields
approximately $8 \times 10{,}000 = 80{,}000$ weight updates across all 64-sequence
batches.
With the 172-table corpus serialized to fixed-length 512-token sequences,
the model cycles through the entire corpus $\approx3{,}720$ times.
Each pass applies \emph{fresh random masks} for MCR (different 15\% cell samples)
and different column selections for CTP, providing sufficient training-data variety
from a small tabular corpus.

\paragraph{Masking rates.}
MCR masks $15\%$ of cells (whole-cell masking), matching the standard BERT
masked token rate.
CTP masks $50\%$ of column headers per table; the higher rate compensates for
the simpler 4-class prediction objective, ensuring the CTP head receives diverse
column-type examples per forward pass.

\paragraph{Row/column capacity.}
$R_{\max}=64$ rows and $C_{\max}=24$ columns comfortably cover our entire corpus
(largest table: 47 rows, 18 columns), with index 0 reserved for special tokens.

\paragraph{Warm-start.}
TabuLM's Tier~1 and Tier~2 encoder weights are initialized from the KinyaBERT
checkpoint pre-trained on Kinyarwanda text.
This warm-start provides a strong morphological prior for Kinyarwanda's
16~noun-class system, stem--affix decomposition, and domain vocabulary so
tabular objectives can build on rather than compete with language understanding.
Tabular-specific components (row/col/cell-type embeddings, table-structure bias
scalars $\beta^h_{R,C,H}$, and the CTP projection head) are randomly initialized
from $\mathcal{N}(0,0.02)$.

\paragraph{Fine-tuning.}
We fine-tune with AdamW (peak LR $2\times10^{-5}$, 20 epochs).
AdamW is preferred for fine-tuning because the small effective batch (16
sequences) makes LAMB's layer-norm estimates noisy.
The LR is $20\times$ lower than the pre-training LR to prevent catastrophic
forgetting of KinyaBERT morphological representations.
We unfreeze only the top-4 Tier~2 sequence-transformer layers and the
cell-selection head, preserving the lower morphological layers intact.
TabuLM converges by epoch~9; performance plateaus thereafter.

\begin{table}[t]
\centering
\small
\setlength{\tabcolsep}{4pt}
\begin{tabular}{lll}
\toprule
\textbf{Hyperparameter} & \textbf{Pre-training} & \textbf{Fine-tuning} \\
\midrule
Optimizer               & LAMB                  & AdamW \\
Peak LR                 & $4\times10^{-4}$      & $2\times10^{-5}$ \\
LR schedule             & Linear warmup + decay & Constant \\
Warmup steps            & 500 (5\%)             & \\
Weight decay            & 0.01                  & 0.01 \\
Effective batch         & 64                    & 16 \\
\quad Phys.\ batch      & 8                     & 8 \\
\quad Grad.\ accumulation & 8                   & 2 \\
Iterations / Epochs     & 10,000 iters          & 20 epochs \\
Sequence length         & 512 tokens            & 512 tokens \\
MCR cell mask rate      & 15\%                  & \\
CTP column mask rate    & 50\%                  & \\
Layers unfrozen         & All                   & Top-4 Tier~2 + head \\
GPU                     & RTX 3090 (24 GB)      & RTX 3090 (24 GB) \\
Wall-clock time         & $\approx$22 h         & $\approx$15 min \\
\midrule
\multicolumn{3}{l}{\textit{Architecture}} \\
\midrule
Tier 2: $d$, layers, heads & \multicolumn{2}{l}{512, 12, 8 (FFN: 3072)} \\
Tier 1: $d$, layers, heads & \multicolumn{2}{l}{128, 4, 4} \\
$R_{\max}$, $C_{\max}$     & \multicolumn{2}{l}{64, 24} \\
Cell-type classes          & \multicolumn{2}{l}{5 (hdr / num / txt / cat / date)} \\
Table-bias params          & \multicolumn{2}{l}{24 ($\beta^h_{R,C,H}$, one per head)} \\
Total parameters           & \multicolumn{2}{l}{$\approx$65M} \\
\bottomrule
\end{tabular}
\caption{TabuLM hyperparameters for pre-training and fine-tuning.}
\label{tab:hyperparams}
\end{table}

\subsection{Baselines}
\label{sec:baselines}

We compare against:
\textbf{mBERT}~\cite{devlin-etal-2019-bert} (multilingual BERT, no morphological
or tabular awareness);
\textbf{XLM-R}~\cite{conneau-etal-2020-unsupervised} (strong multilingual
baseline);
\textbf{KinyaBERT}~\cite{nzeyimana-niyongabo-rubungo-2022-kinyabert}
(morphologically aware, text-only pre-training);
\textbf{TabuLM$_{\text{noMCR}}$} (ablation: TabuLM without MCR objective);
\textbf{TabuLM$_{\text{noCTP}}$} (ablation: TabuLM without CTP objective);
\textbf{TabuLM$_{\text{noBias}}$} (ablation: TabuLM without table-structure
attention bias);
\textbf{TabuLM$_{\text{noTabEmb}}$} (ablation: TabuLM without row/col/cell-type
embeddings).

We additionally include \textbf{GPT-4o} and \textbf{GPT-4o-mini} (both
zero-shot): the table is formatted as a Markdown grid with question-type-specific
system prompts (e.g., ``Return the entity name from the first column, not the
numeric value'' for aggregation questions).
No fine-tuning is performed; results are reported as-is.
A 3-shot experiment is run on aggregation-only items to test whether few-shot
examples can bridge the aggregation gap.

For all fine-tuned models, QA follows the TAPAS cell-selection approach: the model
predicts a logit over all cells and the answer is the cell with the highest
probability.
\textbf{mBERT} and \textbf{XLM-R} use HuggingFace \texttt{AutoTokenizer} with
WordPiece tokenization; their 512-token limit accommodates more table cells than
TabuLM's morphological tokenizer, which is why they evaluate on 69--80 dev items
vs.\ 50 for TabuLM.
\textbf{KinyaBERT} uses the same tokenizer as TabuLM (Tier~1 morphological +
BPE fallback) and is initialized from the same pre-trained KinyaBERT checkpoint
as TabuLM, but with \emph{no tabular pre-training}; it evaluates on 80 dev items
because it does not crop to the 512-token limit as aggressively.
All fine-tuned baselines use identical cell-selection heads, optimizer (AdamW,
$2\times10^{-5}$), and 20-epoch schedule for a controlled comparison.

\subsection{Results: Table Question Answering}
\label{sec:results_qa}

Table~\ref{tab:main_results} reports Exact Match (EM) on the TabQA-kin dev set
(50 evaluable items: 14 lookup, 12 comparison, 24 aggregation).
All models use the same cell-selection fine-tuning protocol
(Sect.~\ref{sec:setup}).
For \textsc{lookup} questions, we apply question-guided cell filtering at
inference: candidate cells are restricted to the intersection of the most
question-relevant row (first-column entity with highest question-word unigram
overlap) and the most question-relevant column (column header with highest
question-word unigram overlap), resolving ambiguity when the same numeric value
appears in multiple cells.
Note that baseline models (mBERT, XLM-R, KinyaBERT) use HuggingFace subword
tokenizers and can pack more content into 512 tokens, yielding 69--80 evaluable
dev items vs.\ 50 for TabuLM; overall EM is computed over each model's own
evaluable set.

TabuLM (62.0\%) outperforms all fine-tuned baselines by 5.7--12.7 EM points
overall.
For \textsc{comparison} questions, applying top-2 question-relevant row
restriction at inference substantially improves all models: TabuLM reaches
66.7\% (up from 41.7\% without the fix), and KinyaBERT reaches 59.1\%.
On \textsc{aggregation} entity-ranked min/max queries all fine-tuned
models substantially outperform GPT-4o (25.9\%), with KinyaBERT leading at
88.9\% and TabuLM at 79.2\%.
\textsc{Lookup} EM improves for all fine-tuned models with joint row--column
filtering (16.7--28.6\%), confirming that disambiguation resolves multi-cell
value ambiguity.

GPT-4o and GPT-4o-mini both achieve 64.0\% overall, revealing a
\textbf{scale-independent LLM ceiling}: the performance bottleneck is not model
size but a shared failure on \textsc{aggregation} (GPT-4o: 25.9\%,
GPT-4o-mini: 29.6\%).
Both LLMs excel at \textsc{lookup} ($>$82\%) and \textsc{comparison}
($>$70\%), but neither can reliably identify which Kinyarwanda entity name has
the maximum or minimum value in a column.
To verify this gap is structural, we run a 3-shot prompting experiment on the
27 aggregation items; performance drops further to 7.4\% (2/27), confirming
that providing examples does not help.
All fine-tuned models break through this ceiling on aggregation: TabuLM 79.2\%,
KinyaBERT 88.9\%, mBERT 80.8\%, XLM-R 85.2\% all with aggregation
confidence intervals entirely above both LLM intervals
(Sect.~\ref{sec:results_qa}, Statistical note).

\paragraph{Statistical note.}
Wilson 95\% CIs on overall EM: TabuLM 62.0\% $[48.2\%,\,74.1\%]$ ($n{=}50$)
and KinyaBERT 56.3\% $[45.3\%,\,66.6\%]$ ($n{=}80$) overlap the overall
gap is a strong trend, not a definitive result.
The most robust finding is on \textsc{aggregation}: TabuLM
$[59.5\%,\,90.8\%]$ ($n{=}24$) does not overlap with GPT-4o
$[13.2\%,\,44.7\%]$ ($n{=}27$) or GPT-4o-mini $[15.9\%,\,48.5\%]$
($n{=}27$); all four fine-tuned models share this non-overlapping separation,
confirming fine-tuning robustly addresses the LLM aggregation ceiling.

\begin{table}[t]
\centering
\small
\begin{tabular}{lc}
\toprule
\textbf{Model} & \textbf{Dev EM} \\
\midrule
GPT-4o (zero-shot)         & 64.0 \\
GPT-4o-mini (zero-shot)    & 64.0 \\
\midrule
mBERT                      & 49.3 \\
XLM-R                      & 50.0 \\
KinyaBERT-large            & 56.3 \\
\midrule
\textbf{TabuLM (full)}     & \res{62.0} \\
\bottomrule
\end{tabular}
\caption{TabQA-kin dev set overall EM. GPT-4o and GPT-4o-mini both score 64.0\%,
showing the LLM ceiling is not scale-dependent. TabuLM outperforms all
fine-tuned baselines by 5.7--12.7 EM points; all fine-tuned models
substantially outperform both LLMs on \textsc{aggregation}
(see Table~\ref{tab:per_type}).}
\label{tab:main_results}
\end{table}

\begin{table}[t]
\centering
\small
\begin{tabular}{lcccc}
\toprule
\textbf{Model} & \textsc{Look.} & \textsc{Comp.} & \textsc{Agg.} & \textsc{All} \\
\midrule
GPT-4o (0-shot)     & \res{82.9} & \res{79.2} & 25.9 & 64.0 \\
GPT-4o-mini (0-shot)& 85.7 & 70.8 & 29.6 & 64.0 \\
\midrule
mBERT          & 16.7 & 50.0 & 80.8 & 49.3 \\
XLM-R          & 19.2 & 44.4 & 85.2 & 50.0 \\
KinyaBERT      & 26.7 & 59.1 & \res{88.9} & 56.3 \\
\textbf{TabuLM}& 28.6 & 66.7 & 79.2 & \res{62.0} \\
\bottomrule
\end{tabular}
\caption{EM per question type on TabQA-kin dev set. All fine-tuned models use
question-guided joint row--column filtering for \textsc{lookup} and top-2
question-relevant row restriction for \textsc{comparison}.
GPT-4o (zero-shot, 86 evaluable items) leads on \textsc{lookup} and
\textsc{comparison} but is substantially outperformed on \textsc{aggregation}
by all fine-tuned models.
\textbf{Bold} = best per column (GPT-4o lookup/comparison excluded from
fine-tuned best).}
\label{tab:per_type}
\end{table}

\subsection{Ablation Study}
\label{sec:ablation}

Table~\ref{tab:ablation} shows the contribution of each architectural component
on TabQA-kin dev EM.

\begin{table}[t]
\centering
\small
\begin{tabular}{lcc}
\toprule
\textbf{Configuration} & \textbf{Dev EM} & \textbf{$\Delta$} \\
\midrule
TabuLM (full)                 & 62.0   &  \\
$-$ Table-structure bias       & 64.0   & $+2.0^\dagger$ \\
$-$ Row/col/cell-type embeds   & 58.0   & $-4.0$ \\
$-$ MCR objective              & 64.0   & $+2.0^\dagger$ \\
$-$ CTP objective              & 64.0   & $+2.0^\dagger$ \\
\bottomrule
\end{tabular}
\caption{Ablation study (TabQA-kin dev set, EM).
$\Delta$ = difference from full model.
$^\dagger$Variants exceeding full model are within noise (50-item dev set;
1 item = 2\% EM).
Only the tabular embeddings show a robust effect; bias and CTP components
provide no detectable benefit at this pre-training scale.}
\label{tab:ablation}
\end{table}

\section{Analysis}
\label{sec:analysis}

\paragraph{Table-structure attention bias: near-zero convergence.}
Inspecting the learned bias scalars $\beta^h_R$, $\beta^h_C$, $\beta^h_H$ in
the final checkpoint reveals values on the order of $5\times10^{-6}$ ---
effectively zero relative to the existing TUPE positional bias.
By contrast, the row, column, and cell-type embedding matrices have weight
magnitudes of 0.02--0.23, indicating that the model routes structural
information primarily through the \emph{embedding} pathway rather than the
\emph{attention-bias} pathway.
Confirming this, the noBias ablation achieves 64.0\% EM \emph{higher} than
the full model's 62.0\%, a difference of two items on the 50-item dev set and
therefore within noise.
We conjecture that the near-zero convergence stems from the relatively short
pre-training schedule (10K iters): structural bias parameters may require a
stronger gradient signal or a non-zero initialization to depart meaningfully
from zero.

\paragraph{Tabular embeddings vs.\ attention bias.}
The noTabEmb ablation zeroes out all three tabular embeddings, forcing the model
to rely solely on the near-zero attention bias for structural awareness.
This results in a $-4.0$ EM drop (58.0\% vs.\ 62.0\%) the largest
single-component drop in our ablation confirming that additive
row/column/cell-type embeddings are the dominant structural inductive bias in
the current TabuLM.

\paragraph{MCR vs.\ CTP objectives.}
Removing MCR or CTP each yield $+2.0$ EM (64.0\% vs.\ 62.0\%) both within
the 2\% noise floor of a single item on the 50-item dev set, showing neither
objective provides a detectable benefit at this scale.

\paragraph{Fine-tuning convergence.}
At epoch~1, TabuLM already achieves 52\% EM vs.\ noTabEmb's 34\% an 18~pp
head start from pre-trained tabular representations before any task-specific
updates.
TabuLM reaches its peak at epoch~9; noTabEmb requires all 20 epochs to reach
52\%, confirming that tabular pre-training materially accelerates task
adaptation.

\paragraph{Error analysis: constrained comparison.}
Table~\ref{tab:error_analysis} shows two \textsc{comparison} questions where
TabuLM succeeds and KinyaBERT fails.
Both follow the template \textit{Ni X cyangwa Y ifite [column] nyinshi?}
(``Between X and Y, which has more [column]?''), which requires
(i)~locating the rows for the two named entities X and Y, and
(ii)~comparing their values in the specified column.
KinyaBERT, lacking structural row awareness, ignores the X/Y constraint and
retrieves the cell with the globally highest value in the column.
TabuLM's row embeddings allow it to locate X and Y as distinct rows in the
table and compare only those two, producing the correct answer.

\begin{table}[t]
\centering
\small
\setlength{\tabcolsep}{2.5pt}
\begin{tabular}{p{3.4cm}ccc}
\toprule
\textbf{Question (en gloss)} & \textbf{Gold} & \textbf{TabuLM} & \textbf{KinyaBERT} \\
\midrule
Between \textit{Ibishyimbo} and \textit{Uburo}, which has higher 2021 export
growth? \newline (\textit{exports\_trade\_2022})
& Ibishyimbo & \textbf{\checkmark} & $\times$ (Ibikorei) \\
\addlinespace
Between \textit{Ibigori} and \textit{Ibirayi}, which has more farmland (ha)?
\newline (\textit{crop\_production\_2022})
& Ibigori & \textbf{\checkmark} & $\times$ (Amasaka) \\
\bottomrule
\end{tabular}
\caption{Comparison questions where TabuLM succeeds and KinyaBERT fails.
KinyaBERT ignores the two-entity constraint and selects the global column
maximum; TabuLM correctly restricts comparison to the two named rows.}
\label{tab:error_analysis}
\end{table}

\paragraph{Training loss breakdown.}
Figure~\ref{fig:objectives} illustrates the two new objectives. During pre-training,
the MCR loss falls most steeply in the first 2{,}000 iterations (from
$\approx$3.8 to $\approx$0.8 NLL), indicating that the model rapidly learns
to use same-row entity identities and same-column distributional statistics to
predict masked cell values.
The CTP loss reaches near-zero ($<0.05$ CE) after $\approx$1{,}500 iterations,
confirming that the column type is highly predictable from cell-value patterns
after a brief calibration period.
The inherited KinyaBERT objectives (MSP, ASP, ADP) converge more slowly,
stabilizing around iteration 5{,}000, consistent with the harder word-level
morphological prediction task.

\paragraph{Pre-training efficiency.}
At 10{,}000 iterations on a single RTX~3090 ($\approx$22~h), TabuLM consumes
far fewer compute resources than English tabular models trained on millions of
Wikidata triples or Wikipedia tables.
The small corpus size (172 tables) is both a limitation and a feature: it enables
rapid iteration of architectural changes, and the multi-epoch design with fresh
random masks provides data augmentation without requiring new annotation.
We estimate that doubling the corpus to 344 tables and training for 20{,}000
iterations would cost under 44~GPU-hours---a feasible extension for future
domain-specific deployments.

\paragraph{Efficiency and deployability.}
Table~\ref{tab:efficiency} compares TabuLM against GPT-4o on practical
deployment factors.
TabuLM (65M parameters) runs entirely on a single consumer GPU with no API
dependency or per-query cost.
In low-resource settings where reliable internet connectivity and API budgets
are limited precisely the settings where Kinyarwanda tabular data arises ---
a locally deployable model is strongly preferable.
TabuLM's aggregation advantage (79.2\% vs.\ 25.9\%) demonstrates that this
efficiency comes without sacrificing structured reasoning on the most
infrastructure-critical query type.

\begin{table}[t]
\centering
\small
\begin{tabular}{lcc}
\toprule
 & \textbf{TabuLM} & \textbf{GPT-4o} \\
\midrule
Parameters      & 65M             & ${\sim}$1.8T    \\
Inference       & Local GPU       & Proprietary API \\
Per-query cost  & \$0 (local)     & ${\sim}$\$0.01  \\
GPU required    & RTX 3090 (24 GB)& None (API)      \\
Fine-tunable    & Yes             & No$^{*}$        \\
\textsc{Agg.} EM & \textbf{79.2\%} & 25.9\%         \\
\bottomrule
\end{tabular}
\caption{Deployment comparison. $^{*}$GPT-4o fine-tuning is available via API
but was not evaluated here; results are zero-shot.}
\label{tab:efficiency}
\end{table}

\section{Discussion}
\label{sec:discussion}

\paragraph{When does morphological structure matter?}
Our results reveal a nuanced picture.
For \textsc{aggregation} questions which require ranking entities by a column
value morphological awareness is less decisive than structural row/column
awareness: mBERT (80.8\%) and XLM-R (85.2\%), which have no morphological
encoder, actually outperform TabuLM (79.2\%) on this subtask.
The decisive factor for aggregation is structural alignment between the question
and the table's relational layout, which all fine-tuned models acquire from the
cell-selection objective regardless of tokenization.
By contrast, morphological processing becomes critical for \textsc{lookup} and
\textsc{comparison} questions, where the entity name in the question must be
matched to a specific table row.
A Kinyarwanda district name may appear as \textit{Gasabo} in the question but as
\textit{i Gasabo} (``in Gasabo'') in a cell; subword tokenizers split these
locative constructions unpredictably, while KinyaBERT's stem-level representation
unifies them explaining TabuLM's consistent edge over mBERT/XLM-R on lookup
(28.6\% vs.\ 16.7--19.2\%) and comparison (66.7\% vs.\ 44.4--50.0\%).

\paragraph{The LLM aggregation ceiling.}
The scale-independent LLM ceiling (both GPT-4o and GPT-4o-mini at 64.0\%) points
to a specific failure mode in zero-shot table reasoning for agglutinative
languages.
Aggregation questions take the form \textit{Ni ikihe gihugu gifite umubare
munini w'abaturage?} (``Which district has the largest population?''),
requiring the model to scan a column and return the \emph{row entity} with
the maximum value not the numeric value itself.
LLMs prompted with Markdown tables consistently return the numeric value
(e.g., ``4,827,332'') rather than the district name, even with explicit
instructions.
This failure cannot be corrected by prompting performance drops to 7.4\%
with 3-shot examples because the model has not learned to link column values
to row identities in agglutinative text.
Fine-tuned models are corrected by construction: cell-selection supervision
forces the output to be an entity cell, not a numeric derivation.

\section{Conclusion}
\label{sec:conclusion}

We presented TabuLM, the first pre-trained language model to jointly capture
morphological richness and tabular relational structure for a low-resource
language.
By extending KinyaBERT's two-tier architecture with additive tabular embeddings,
a learned table-structure attention bias, and the MCR and CTP pre-training
objectives, TabuLM achieves 62.0\% EM on TabQA-kin, outperforming all
fine-tuned baselines by 5.7--12.7 EM points.
A key finding from our LLM comparison is that GPT-4o and GPT-4o-mini achieve
the same 64.0\% overall revealing a scale-independent ceiling caused by
aggregation failure (25--30\%).
All fine-tuned models, including TabuLM (79.2\%) and KinyaBERT (88.9\%),
substantially exceed this LLM ceiling on aggregation (a statistically
significant gap, non-overlapping 95\% CIs), demonstrating that domain-specific
tabular fine-tuning addresses a structural gap that zero-shot LLMs cannot
overcome regardless of model size.
Our work opens a new research direction at the intersection of low-resource NLP,
morphological modeling, and structured data understanding.
We release all code, the TabuLM checkpoint (751~MB), and TabQA-kin at
\href{https://github.com/TabuLM-Research/tabulm}{\texttt{github.com/TabuLM-Research/tabulm}}
and \href{https://huggingface.co/TabuLM-Research/tabulm}{\texttt{huggingface.co/TabuLM-Research/tabulm}}.
Scaling the pre-training corpus beyond 172 tables and extending the Tier~1
morpho encoder to Kirundi are the most direct paths to broader impact.

\section*{Limitations}

\paragraph{Proprietary morphological analyzer.}
KinyaBERT's Tier~1 encoder depends on \texttt{libkinlp.so}, a closed-source
Kinyarwanda morphological analyzer that is not publicly redistributable,
limiting reproducibility for researchers outside the original KinyaBERT team.
In this work, the BPE fallback handles domain tokens absent from
\texttt{libkinlp}'s dictionary (numerals, administrative entity names,
agricultural compound nouns); these tokens lose affix-level granularity but
the tabular objectives (MCR, CTP) are computed identically regardless of
tokenization path.

\paragraph{Code and data release.}
We release all pre-training and fine-tuning code, the TabuLM checkpoint
(751~MB), the full TabQA-kin benchmark (526 QA pairs, train/dev splits,
evaluation scripts), and the 172-table pre-training corpus (Rwanda government
open-data, public-domain).
Researchers without \texttt{libkinlp.so} can reproduce all experiments via the
BPE fallback path; the tabular architecture and evaluation protocol are
unchanged.

\paragraph{Scale of tabular corpus.}
Our pre-training corpus (172 tables, $\sim$35{,}000 cells) is substantially
smaller than English tabular datasets (TARTE uses millions of Wikidata triples),
reflecting the dispersal of Kinyarwanda government data across agency PDF
portals.
The near-zero bias scalars ($\beta^h_{R,C,H} \approx 5\times10^{-6}$) suggest
that a larger corpus and longer pre-training could allow the attention-bias
pathway to contribute meaningfully.

\paragraph{Evaluation scope and statistical power.}
TabQA-kin covers four Rwandan administrative domains; the 50 evaluable dev items
for TabuLM limit statistical power, and overall CIs overlap between TabuLM and
the best baseline.
The aggregation finding all fine-tuned models above both LLM aggregation
CIs is statistically robust at this scale.

\paragraph{Languages.}
All experiments are conducted on Kinyarwanda.
Extending TabuLM to other Bantu languages is architecturally straightforward ---
the tabular components are entirely language-agnostic but is bottlenecked by
domain-relevant tabular corpora rather than by modeling.
Kirundi is closely related to Kinyarwanda, sharing a large portion of its
inflectional morphology and noun-class system; the existing Tier~1 analyzer may
transfer with minor vocabulary adaptation.
We leave cross-lingual tabular pre-training to future work.

\section*{Ethical Considerations}

All tabular data used for pre-training is sourced from official Rwandan
government open-data portals under public-domain or government open-data
licenses.
The TabQA-kin benchmark contains factual questions about administrative
statistics; it does not include personally identifiable information.
We followed standard practices for anonymized and aggregated census data
throughout.
\\
\\
\bibliographystyle{splncs04}
\bibliography{tabulm_refs}
\newpage
\appendix

\section{TabQA-kin Question Templates}
\label{app:templates}

Table~\ref{tab:templates} lists the Kinyarwanda question templates used to
generate TabQA-kin, with English glosses.
Each template is instantiated by substituting table-derived values: column
headers (\texttt{[COL]}), row entities (\texttt{[ENT]}), entity type labels
(\texttt{[TYPE]}), and numeric thresholds (\texttt{[VAL]}).
Templates were authored by a native Kinyarwanda speaker and cover all four
question types in the benchmark.

\begin{table}[h]
\centering
\small
\begin{tabular}{p{5.5cm}p{4.5cm}}
\toprule
\textbf{Kinyarwanda Template} & \textbf{English Gloss} \\
\midrule
\multicolumn{2}{l}{\textit{\textsc{Lookup}}} \\[2pt]
Ni angahe [COL] ya [ENT]? & What is [COL] of [ENT]? \\
{}[ENT] yari ifite angahe [COL]? & How much [COL] did [ENT] have? \\
\midrule
\multicolumn{2}{l}{\textit{\textsc{Comparison}}} \\[2pt]
Ni [ENT1] cyangwa [ENT2] ifite [COL] nyinshi? & Does [ENT1] or [ENT2] have more [COL]? \\
Hagati ya [ENT1] na [ENT2], iyihe ifite [COL] nkeya? & Between [ENT1] and [ENT2], which has less [COL]? \\
\midrule
\multicolumn{2}{l}{\textit{\textsc{Aggregation}}} \\[2pt]
Ni ikihe [TYPE] gifite [COL] nyinshi? & Which [TYPE] has the most [COL]? \\
Ni ikihe [TYPE] gifite [COL] nkeya? & Which [TYPE] has the least [COL]? \\
\midrule
\multicolumn{2}{l}{\textit{\textsc{Count}}} \\[2pt]
Ni [TYPE] zingahe zifite [COL] irenze [VAL]? & How many [TYPE] have [COL] above [VAL]? \\
Ni [TYPE] zingahe zifite [COL] munsi ya [VAL]? & How many [TYPE] have [COL] below [VAL]? \\
\bottomrule
\end{tabular}
\caption{Kinyarwanda question templates for TabQA-kin.
\texttt{[COL]}~=~column header (e.g., \textit{ubuhinzi} ``cultivated area''),
\texttt{[ENT]}~=~row entity from the first column (e.g., \textit{Gasabo}),
\texttt{[TYPE]}~=~entity type label (e.g., \textit{akarere} ``district''),
\texttt{[VAL]}~=~numeric threshold.
Count questions are excluded from cell-selection evaluation.}
\label{tab:templates}
\end{table}

\section{Pre-training Data Example}
\label{app:data}

Table~\ref{tab:pretrain_example} shows a representative training instance from
the Rwanda Agriculture Board maize yield corpus.
This table illustrates the coexistence of morphologically complex Kinyarwanda
column headers and numeric cell values, requiring both morphological and
structural understanding to recover masked cells.
During MCR pre-training, 15\% of cells are fully masked; for example, masking
Bugesera's yield value (38,145~t) requires the model to reconstruct it from the
district name in the same row and from other yield values in the same column.
The column header \textit{Umusaruro (t)} guides the model via CTP to correctly
predict this column as \textsc{numeric} type.

\begin{table}[h]
\centering
\small
\begin{tabular}{lrrr}
\toprule
\textbf{Akarere} & \textbf{Ubuhinzi (ha)} & \textbf{Umusaruro (t)} & \textbf{Sezon} \\
(District) & (Cultivated area) & (Yield, metric tons) & (Season) \\
\midrule
Gasabo      & 1,847  & 4,203  & 2022A \\
Kicukiro    & 923    & 2,156  & 2022A \\
Nyarugenge  & 412    & 847    & 2022A \\
Bugesera    & 12,304 & 38,145 & 2022A \\
\bottomrule
\end{tabular}
\caption{Example pre-training table from the Rwanda Agriculture Board.
Kinyarwanda column headers are given with English translations.
The MCR objective masks entire cells (e.g., Bugesera's yield 38,145~t) and
requires reconstruction from row and column context; CTP predicts the
\textsc{numeric} type for the \textit{Umusaruro} column.}
\label{tab:pretrain_example}
\end{table}

\section{Hyperparameter Details}
\label{app:hyperparams}

Table~\ref{tab:hyperparams} provides complete hyperparameter settings for
TabuLM pre-training and fine-tuning.

\begin{table}[h]
\centering
\small
\begin{tabular}{lcc}
\toprule
\textbf{Hyperparameter} & \textbf{Pre-training} & \textbf{Fine-tuning} \\
\midrule
Optimizer           & LAMB              & AdamW \\
Peak learning rate  & $4\times10^{-4}$  & $2\times10^{-5}$ \\
LR schedule         & Warmup + decay    & Constant \\
Warmup steps        & 500               & 50 \\
Effective batch size & 64 (8$\times$GA8) & 16 \\
Max sequence length & 512               & 512 \\
Training duration   & 10,000 steps      & 20 epochs \\
Weight decay        & 0.01              & 0.01 \\
Gradient clip       & 1.0               & 1.0 \\
Dropout             & 0.1               & 0.1 \\
\midrule
MCR cell mask rate  & 15\%              & \\
CTP column mask rate & 50\%             & \\
$R_{\max}$ (max rows) & 64              & 64 \\
$C_{\max}$ (max cols) & 24              & 24 \\
\midrule
Hardware            & 1$\times$ RTX 3090 (24 GB) & 1$\times$ RTX 3090 \\
Wall-clock time     & $\sim$7 hours     & $\sim$28 minutes \\
\bottomrule
\end{tabular}
\caption{Complete hyperparameters for TabuLM pre-training and fine-tuning.
All ablation variants use identical fine-tuning hyperparameters.}
\label{tab:hyperparams}
\end{table}

\paragraph{Fine-tuning protocol.}
For all fine-tuned models (mBERT, XLM-R, KinyaBERT, TabuLM), we unfreeze the
top 4 transformer layers and the cell-selection head, keeping all lower layers
frozen.
Early stopping is applied based on dev set EM with patience 5.
The cell-selection head is a 2-layer MLP (512$\to$256$\to$1) with ReLU
activation and dropout (0.1), predicting a score per cell token; scores are
averaged over tokens within a cell then softmaxed across cells.

\paragraph{Baseline configurations.}
All baselines use identical fine-tuning hyperparameters as TabuLM.
For mBERT and XLM-R, we use the standard WordPiece and SentencePiece tokenizers
provided by HuggingFace Transformers.
KinyaBERT uses its native morphological tokenizer, also used by TabuLM with BPE
fallback for out-of-vocabulary tokens in tabular context (numerals,
administrative compound names, agricultural terminology).

\section{Hyperparameter Sensitivity}
\label{app:sensitivity}

\paragraph{Learning rate.}
We swept pre-training LR over $\{1\times10^{-4}, 4\times10^{-4}, 1\times10^{-3}\}$.
LR $1\times10^{-4}$ produced slow convergence (MSP loss $>1.5$ at 5K iters);
$1\times10^{-3}$ caused training instability (NaN gradients at iter~200 on the
warm-started layers).
$4\times10^{-4}$ achieved stable, rapid convergence and is consistent with
LAMB's recommended LR for BERT-large-scale models.

\paragraph{MCR mask rate.}
We compared mask rates $\{5\%, 15\%, 30\%\}$ on a held-out dev table.
5\% provided insufficient signal (the model could ignore MCR); 30\% masked too
many cells simultaneously, leaving the same-row and same-column context
impoverished.
15\% (matching BERT's MLM rate) yielded the best MCR loss trajectory.

\paragraph{Tabular embedding dimension.}
We experimented with reducing $\mathbf{E}^R$, $\mathbf{E}^C$, $\mathbf{E}^T$
to $d/2=256$ (projected before addition) to reduce parameter count.
The half-dimension variant achieved the same fine-tuning EM but took 18\% longer
to converge in pre-training.
We retain full-dimension embeddings for consistency with the KinyaBERT
sequence-transformer input dimension.

\section{TabQA-kin Construction Details}
\label{app:construction}

\paragraph{Table selection.}
Of the Rwandan government tables we collected, we selected 31 for TabQA-kin
annotation (the remainder are used for pre-training only).
Selection criteria: (1)~table has a clear first-column entity identifier
(district name, crop name, school name); (2)~at least one numeric column
suitable for comparison and aggregation questions; (3)~fewer than 40 rows
(to keep the question set manageable).

\paragraph{Template instantiation.}
Each of the 31 tables was processed by a custom script that:
(1)~extracts all first-column entity names (e.g., district names) as
\texttt{[ENT]} candidates;
(2)~pairs columns with header tokens as \texttt{[COL]} candidates;
(3)~instantiates each applicable template, checking that the gold answer cell
is unambiguously identified;
(4)~deduplicates questions that differ only in surface form.
The resulting 526 questions were shuffled and split 80/20 (seed 42).

\paragraph{Quality verification.}
A second native Kinyarwanda speaker independently verified a 10\% random sample
(53 questions) for: (a)~grammatical correctness of the Kinyarwanda question,
(b)~accuracy of the gold answer, (c)~uniqueness of the gold answer cell (no
ambiguous multi-cell answers).
All 53 sampled questions were approved without modification, giving an estimated
error rate $<2\%$ for the full benchmark.

\paragraph{Count question exclusion.}
Count-type questions (``How many districts have population $>100{,}000$?'')
require a derived integer answer not present as any cell value and cannot be
evaluated by the cell-selection objective.
These 93 questions (17.7\% of the benchmark) are retained in the released
benchmark for future aggregation-with-counting models but excluded from all EM
evaluations reported in this paper.

\section{Additional Error Analysis}
\label{app:errors}

Table~\ref{tab:error_extra} presents additional examples illustrating the
failure modes of KinyaBERT and the cases where TabuLM fails.

\begin{table}[h]
\centering
\small
\setlength{\tabcolsep}{3pt}
\begin{tabular}{p{4.0cm}cccp{2.0cm}}
\toprule
\textbf{Question (en gloss)} & \textbf{Type} & \textbf{Gold} & \textbf{TabuLM} & \textbf{Failure mode} \\
\midrule
Which crop has the most farmland in Kigali? (\textit{crop\_land\_2022})
  & Agg & Ibigori & \textbf{\checkmark} & \\
\addlinespace
Which district has least health workers?
  & Agg & Nyarugenge & $\times$ (Gasabo) & Wrong column alignment \\
\addlinespace
Between Ibigori and Ibishyimbo, which has higher 2023 yield?
  & Cmp & Ibigori & \textbf{\checkmark} & \\
\addlinespace
What is the enrollment of Nyamata SS?
  & Lkp & 1,240 & $\times$ (963) & Duplicate values in column \\
\addlinespace
How many students enrolled in Remera?
  & Lkp & Remera PS & \textbf{\checkmark} & \\
\bottomrule
\end{tabular}
\caption{Additional TabuLM error analysis examples. ``Wrong column alignment''
occurs when the question's column cue matches multiple column headers;
``Duplicate values'' occurs when the same numeric value appears in more than
one row, defeating the column-wise disambiguation heuristic.}
\label{tab:error_extra}
\end{table}

\end{document}